\documentclass[
]{ceurart}

\usepackage{listings}
\usepackage{amsmath, amssymb}
\usepackage{algorithm}
\usepackage{algpseudocode}
\begin{document}

\copyrightyear{2025}
\copyrightclause{Copyright for this paper by its authors.
  Use permitted under Creative Commons License Attribution 4.0
  International (CC BY 4.0).}

\conference{CLEF 2025 Working Notes, 9 -- 12 September 2025, Madrid, Spain}

\title{Tighnari v2: Mitigating Label Noise and Distribution Shift in Multimodal Plant Distribution Prediction via Mixture of Experts and Weakly Supervised Learning}

\title[mode=sub]{Notebook for the <LifeCLEF> Lab at CLEF 2025}

\tnotetext[1]{You can use this document as the template for preparing your
  publication. We recommend using the latest version of the ceurart style.}


\author[1,2]{Haixu Liu}[%
  orcid=0009-0007-8115-0826,
  email=hliu2490@uni.sydney.edu.au，
]
\cormark[1]
\fnmark[1]

\author[2]{Yufei Wang}[%
  orcid=0009-0008-6002-3729,
  email=z5536297@ad.unsw.edu.au,
]
\cormark[1]
\fnmark[1]

\author[3]{Tianxiang Xu}[%
  orcid=0000-0002-6121-2432,
  email=xtx_pku@stu.pku.edu.cn,
]
\cormark[1]
\fnmark[1]

\author[1]{Chuancheng Shi}[%
  orcid=0009-0002-2278-2341,
  email=cshi0459@uni.sydney.edu.au,
]
\cormark[1]
\fnmark[1]

\author[4]{Hongsheng Xing}[%
  orcid=0000-0002-4901-4863,
  email=starxsky@163.com，
]
\cormark[1]
\fnmark[1]

\address[1]{The University of Sydney, Sydney, New South Wales, Australia}

\address[2]{The University of New South Wales, Sydney, New South Wales, Australia}

\address[3]{School of Software and Microelectronics, Peking University, Beijing, China}

\address[4]{Shandong University of Technology, Zibo, Shandong, China}

\cortext[1]{Corresponding author.}
\fntext[1]{These authors contributed equally.}

\begin{abstract}
Large-scale, cross-species plant distribution prediction plays a crucial role in biodiversity conservation, yet modeling efforts in this area still face significant challenges due to the sparsity and bias of observational data. Presence–Absence (PA) data provide accurate and noise-free labels, but are costly to obtain and limited in quantity; Presence-Only (PO) data, by contrast, offer broad spatial coverage and rich spatiotemporal distribution, but suffer from severe label noise in negative samples. To address these real-world constraints, this paper proposes a multimodal fusion framework that fully leverages the strengths of both PA and PO data. We introduce an innovative pseudo-label aggregation strategy for PO data based on the geographic coverage of satellite imagery, enabling geographic alignment between the label space and remote sensing feature space. In terms of model architecture, we adopt Swin Transformer Base as the backbone for satellite imagery, utilize the TabM network for tabular feature extraction, retain the Temporal Swin Transformer for time-series modeling, and employ a stackable serial tri-modal cross-attention mechanism to optimize the fusion of heterogeneous modalities. Furthermore, empirical analysis reveals significant geographic distribution shifts between PA training and test samples, and models trained by directly mixing PO and PA data tend to experience performance degradation due to label noise in PO data. To address this, we draw on the mixture-of-experts paradigm: test samples are partitioned according to their spatial proximity to PA samples, and different models trained on distinct datasets are used for inference and post-processing within each partition. Experiments on the GeoLifeCLEF 2025 dataset demonstrate that our approach achieves superior predictive performance in scenarios with limited PA coverage and pronounced distribution shifts, ranking third in GeoLifeCLEF 2025 (where PA test samples exhibit geographic out-of-distribution characteristics), and surpassing the 2nd-place score on the GeoLifeCLEF 2024 leaderboard (where PA test and training samples are largely identically distributed).

\end{abstract}

\begin{keywords}
Weakly Supervised Learning \sep
Mixture of Experts \sep
Temporal Swin-Transformer \sep
Species Distribution Model 
\end{keywords}

\maketitle

\section{Introduction}
\subsection{Background}
The task of predicting plant species distributions based on spatial location typically involves, at a given set of spatiotemporal coordinates, using environmental, remote‐sensing, and neighborhood features to predict whether a particular plant is likely to occur. This approach enables managers to rapidly identify priority conservation areas at a macro scale and to assess the expansion or contraction of suitable habitat under climate change. However, several challenges remain: observation data are sparse and biased toward hotspot regions and common species; Presence‐Only (PO) data have a much higher error ceiling than Presence–Absence (PA) data; and interspecific interactions are still difficult to model explicitly.

% OK

In the current dataset, we have 88,987 PA training samples, 3,845,533 PO training samples (based on SurveyID merged from 5,079,797 observation records), and 14,716 test samples. Each sample represents a single survey, and its label consists of the identifiers of all plant species observed during that survey, covering a total of 11,255 species 
\cite{geolifeclef2025,lifeclef2025}.
Survey samples are classified as Presence–Absence (PA) or Presence‐Only (PO) depending on whether negative cases strictly represent species that were absent. In real‐world scenarios, PA data originate from plot surveys conducted by official or professional institutions: any species that was ever observed within the survey area is labeled as a positive sample, and any species not observed is explicitly labeled as a negative sample, so there is effectively no label noise. However, because professional researchers are limited in number, while PA labels are cleaner and more accurate, the cost of obtaining them is high, and the total number of PA records is far smaller than that of PO records.
By contrast, PO data typically come from crowdsourced citizen‐science efforts. They are very large in volume and span long time periods, and they provide invaluable spatiotemporal coverage—especially for rare or hard‐to‐monitor species. However, citizens tend to record only species that interest them, while ignoring species they consider “common” or simply do not recognize. As a result, in PO data only positive samples can be taken as fully reliable; negative samples can contain substantial label noise. Therefore, when building species distribution models, PA and PO data each contribute complementary strengths.

Previous work integrating PA and PO data has generally followed one of two paradigms. Chen et al.\cite{chen2024combining} proposed a three‐stage training framework: first, train a network on high‐quality PA data; second, use that network to generate pseudo‐labels for PO data and perform semi‐supervised fine‐tuning; and finally, refine the model again using the PA data. This approach preserves the “data dividend” of PO while mitigating its label noise and distribution shift problems. Liu et al.\cite{liu2025tighnari} proposed an alternative framework based exclusively on PA data: they train a network on PA data and, under the ecological prior that geographic proximity implies ecological similarity, extract the most frequently occurring species among neighboring PO and PA nodes for each test sample to supplement the neural network’s predictions. These two approaches placed second and third, respectively, in the 2024 GeoLifeCLEF challenge with only a small margin between them \cite{picek2024overview}.
\subsection{Our method}

To leverage PO data, we propose a novel pseudo-labeling strategy based on aggregating plant species labels from PO survey samples within the geographic coverage of satellite image patches. Our network is an improved version of the Tighnari model introduced in 2024. With the expansion of training data, we scale up the backbone Swin Transformer for satellite image feature extraction to the Base size, and adopt the TabM network \cite{gorishniy2024tabm}---proposed by Yandex---as the backbone for the tabular modality to enhance feature representation. The Temporal Swin-Transformer for time series feature extraction is retained. Additionally, the modality derived from neighborhood label aggregation is now set as an optional input rather than a mandatory one, as severe label noise in PO data could propagate into the feature space through label aggregation; thus, this modality is only utilized when all training data are from PA samples.

Finally, we improve the modality fusion module by replacing the original hierarchical cross-attention mechanism with a stackable serial tri-modal cross-attention, allowing better fusion of heterogeneous modalities.

During exploratory data analysis (EDA), we discovered distributional differences between the geographic locations of PA test and PA training samples. According to the ecological prior that geographic proximity implies ecological similarity, such distribution shifts can significantly degrade model performance on test samples located far from the PA training distribution. Moreover, models jointly trained on re-labeled PO data (hereafter referred to as PO) and PA data suffer from label noise in the PO data, resulting in inferior inference performance on test samples overlapping with the training distribution compared to models trained solely on PA data. To address this, inspired by the expert model paradigm, we partition test samples based on whether a PA sample exists within a 10-kilometer radius, and assign different models trained on distinct datasets to infer the two parts separately.

Our contributions are as follows:
\begin{enumerate}
    \item We propose a weakly supervised pseudo-labeling rule, aggregating PO sample labels within the geographic coverage of satellite image patches. This reduces label sparsity and minimizes negative label noise without introducing positive label noise.
    \item We design a stackable cross-attention module with serial updates and shared Key/Value layers across modalities, enhancing multimodal fusion capability.
    \item We develop an efficient two-stage training procedure that allows the model to maintain accurate discrimination of positive and negative labels under the PA training sample distribution while retaining the recognition of positive labels learned from the large-scale PO data.
    \item Following the Mixture of Experts (MoE) paradigm, we train separate models on different datasets to predict samples from distinct geographic distributions.
\end{enumerate}

\section{Exploratory Data Analysis }

\subsection{Geographical Distribution of PO and PA Samples}

Exploratory data analysis (EDA) is crucial for uncovering the motivation behind our modeling choices. We visualized the geographic locations of the Survey IDs for both PA (Presence-Absence) and PO (Presence-Only) samples, aiming to investigate their spatial distribution patterns.

\begin{figure}[h]
	\centering
	\includegraphics[width=1\linewidth]{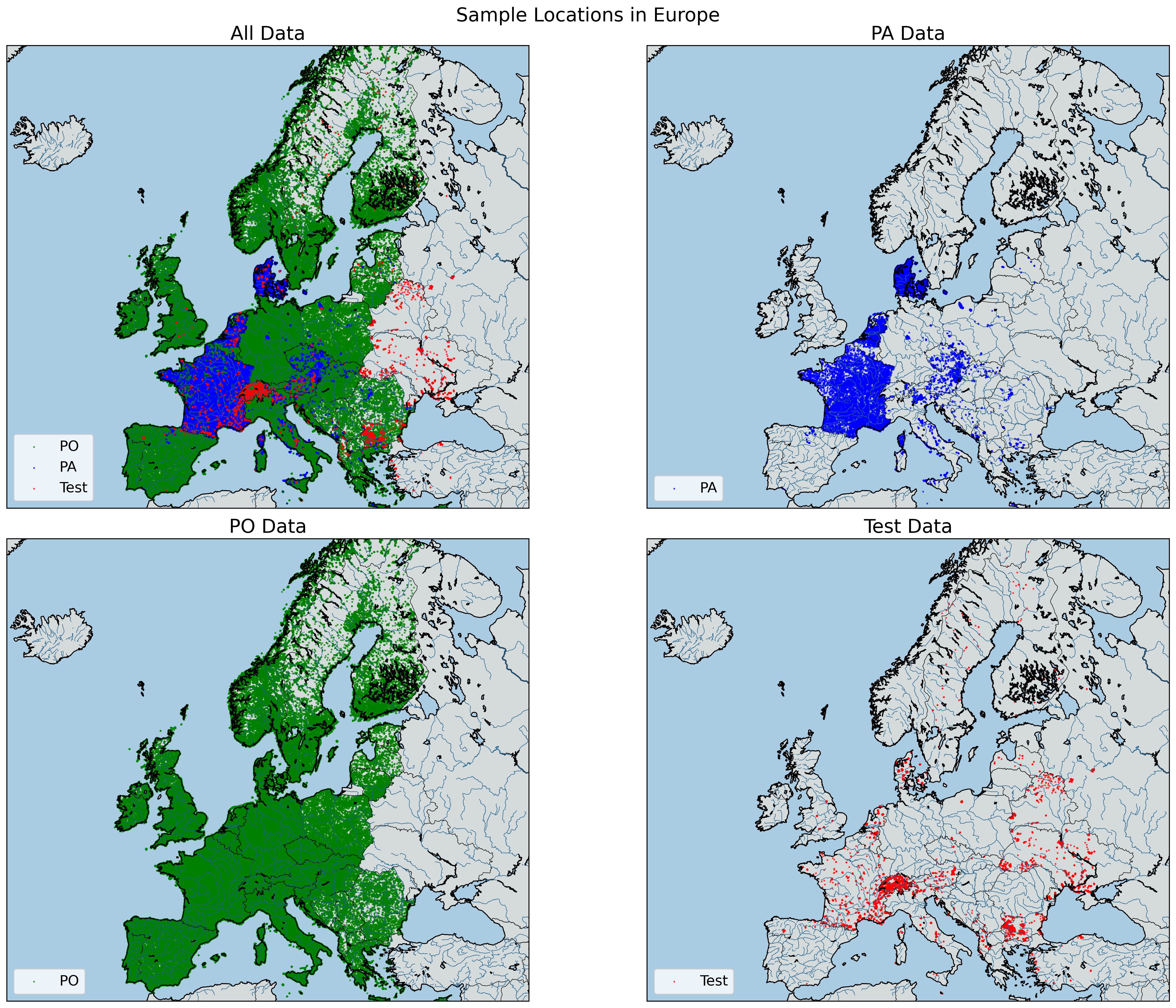}
	\caption{Visualization of survey occurrence locations in the  PO,PA training set as well as test set}
	\label{fig5}
\end{figure}

The visualization reveals significant differences in the geographic distributions among PA training samples, PO training samples, and PA test samples. Specifically, PA training samples are only distributed in Western European countries such as France, the Netherlands, and Denmark. In contrast, PO training samples cover both Western and Central Europe. The PA test samples, in addition to the regions covered by both PA and PO training data, also include certain Eastern European countries such as Ukraine. This indicates the existence of an out-of-distribution (OOD) scenario in the test set  (shown in Figure \ref{fig5}).

\newpage
\subsection{Differences in Species Occurrence Frequency and Sample Species Counts between PO and PA}
\begin{figure}[h]
	\centering
	\vspace{-0.8em}
    \includegraphics[width=1\linewidth]{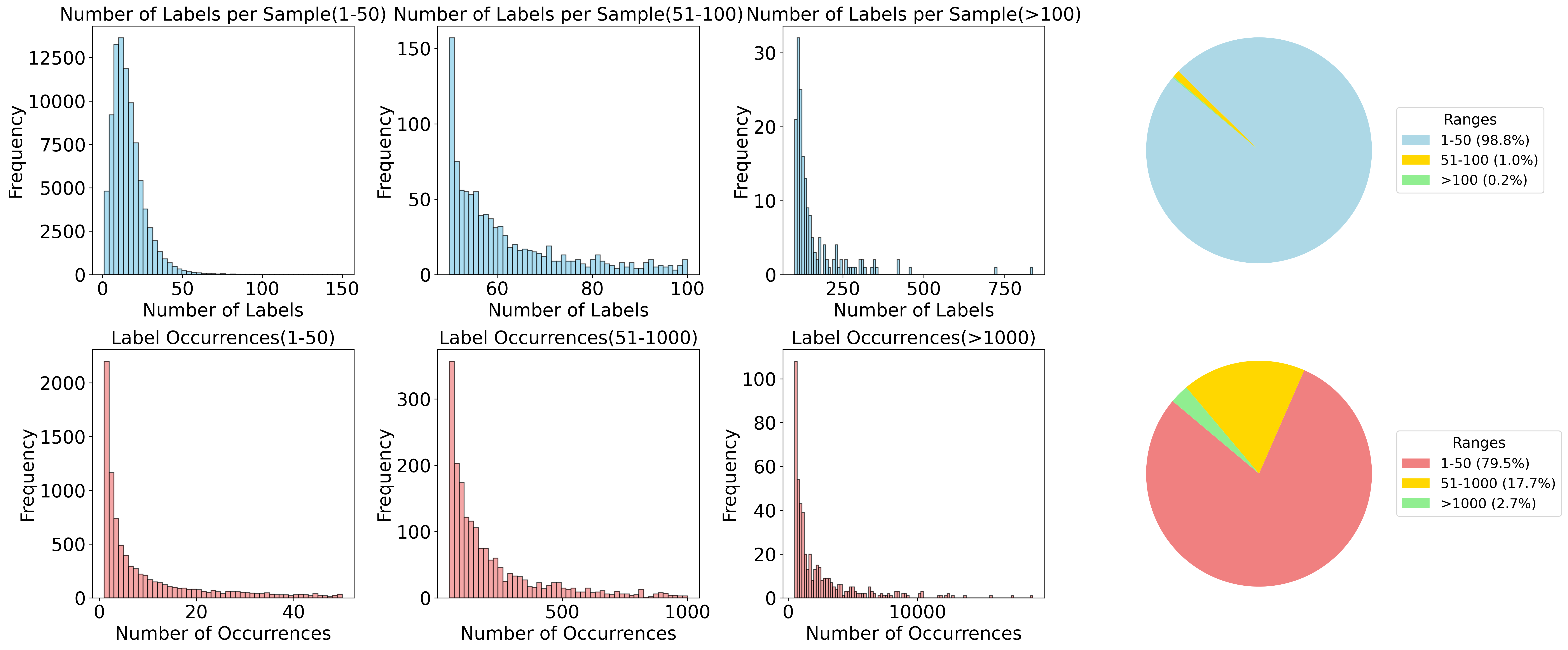}\vspace{-1.5em}
	\caption{Visualization of species occurrence frequency and sample species counts in PA}
	\label{fig1}
    \vspace{-0.8em}
\end{figure}

\begin{figure}[h]
	\centering
	\vspace{-0.8em}
    \includegraphics[width=1\linewidth]{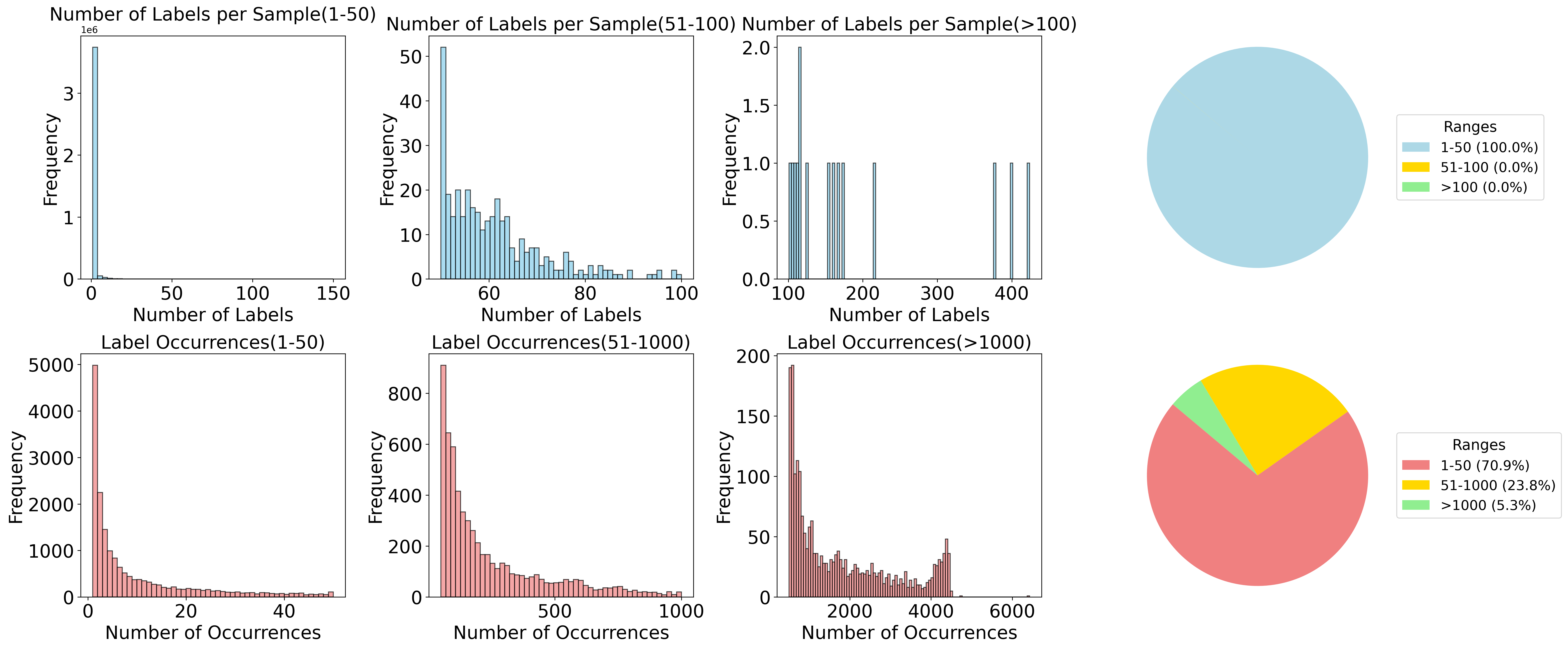}\vspace{-1.5em}
	\caption{Visualization of species occurrence frequency and sample species counts in PO}
	\label{fig2}
    \vspace{-0.8em}
\end{figure}

\begin{figure}[h]
	\centering
	\vspace{-0.8em}
    \includegraphics[width=1\linewidth]{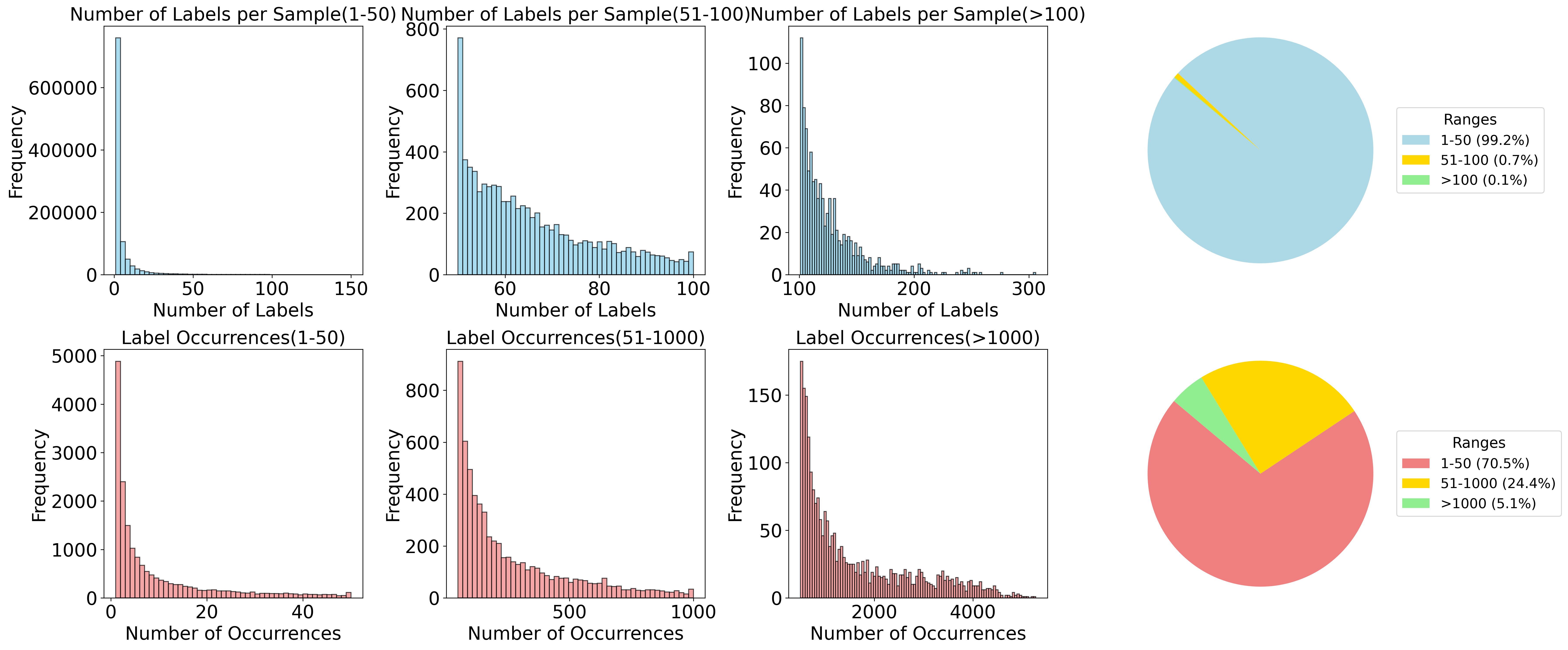}\vspace{-1.5em}
	\caption{Visualization of species occurrence frequency and sample species counts in pseudo-labels of PO}
	\label{fig3}
    \vspace{-0.8em}
\end{figure}

\noindent We plotted three figures to illustrate the presence–absence (PA) dataset (shown in Figure \ref{fig1}), the presence-only (PO) dataset obtained by merging observation records solely by \texttt{SurveyID} (shown in Figure \ref{fig2}), and the PO dataset produced after applying our processing strategy (shown in Figure \ref{fig3}).  
For each dataset, the figures display \textit{(i)} the distribution of the total number of species recorded in each survey and \textit{(ii)} the distribution of the number of surveys in which each species appears.  
A substantial discrepancy between the PA and PO datasets in the per-survey species-count distributions indicates severe label noise, whereas an excessive divergence in the species-frequency distributions implies that the PO samples suffer from a long-tail problem inconsistent with the true ecological situation.

The blue histograms represent the distribution of the number of species present in each survey. It can be observed that the peak of the PA training samples is at 10, while for the unprocessed PO training data, the peak is at 1, and over 99\% of the samples contain fewer than three species. This suggests that the raw PO data contains a large amount of noise. The third figure shows the distribution after relabeling the PO data with our pseudo-labeling strategy. Although the peak is still at 1, the distribution becomes less steep and is closer to the distribution of PA samples.

The red histograms represent the number of occurrences for each species across all surveys. We found that 79.5\% of plant species in the 88,987 PA samples occur fewer than 50 times, while more than 70\% of species in the unprocessed PO samples appear fewer than 50 times, with 5,000 species appearing only once. Although the final evaluation metric is the weighted F1 score, accurately predicting these rare species is often the most critical aspect of species distribution modeling based on spatial and positional data.

\section{Method}
In this section, we present a comprehensive description of the entire process from data preprocessing to model development and training.

\subsection{Pseudo-Labeling Strategy}
We observed significant differences in the annotation procedures between PA (Presence-Absence) and PO (Presence-Only) survey samples. The labels for PA samples are derived from quadrat surveys, in which researchers lay out plots (quadrats) of fixed area and shape—typically square, rectangular, or circular—within the study region and conduct detailed investigations and recordings of the species, abundance, and other attributes within each quadrat. In contrast, PO samples are mainly sourced from citizen science crowdsourcing platforms, where volunteers photograph observed species and upload their geotagged observations. Consequently, PO labels are much sparser compared to PA labels.

If we aggregate (i.e., merge and deduplicate) all PO survey sample labels within a fixed geographic area, it is equivalent to a group of citizen scientists collaboratively conducting a less professional quadrat survey within that area. Thus, the labeling strategy for aggregated PO surveys becomes analogous to that of PA surveys.

Moreover, each survey corresponds to a satellite image patch—a $64 \times 64$ grid of pixels, each with a spatial resolution of $10\,\mathrm{m} \times 10\,\mathrm{m}$—centered on the survey site, forming a $640\,\mathrm{m} \times 640\,\mathrm{m}$ square region. For any PO sample (hereafter referred to as the primary sample for aggregation), if we aggregate the labels of all PO samples located within its corresponding satellite image patch (the $640\,\mathrm{m} \times 640\,\mathrm{m}$ area), the resulting patch effectively contains the satellite image pixels of all aggregated samples, thus preventing the introduction of image feature noise that could arise when the satellite pixels of aggregated PO samples fall outside the patch of the primary sample. Compared to training a teacher model with PA data and then using it to pseudo-label PO samples, this aggregation-based strategy significantly reduces negative label noise without introducing positive label noise.

Another critical issue is that the initial PO dataset comprises 3,845,533 surveys. If every survey were to be pseudo-labeled using this aggregation strategy, the distribution of species occurrences would become even more imbalanced. Conversely, if all PO samples that participate in label aggregation are subsequently excluded from serving as primary samples for further aggregation, the occurrences of rare species would remain scarce. Worse still, since some surveys containing rare species are geographically adjacent, their labels may be merged into a single PO sample, further decreasing their occurrence count.

To address this, we propose a reserved-sample strategy: if a PO survey sample does not contain any species whose overall occurrence count across all surveys is less than 100, then this sample will no longer serve as a primary sample for subsequent aggregation. Otherwise, it remains eligible to aggregate labels from nearby PO samples. We designate the three filtering strategies for PO survey samples as loose filtering, balanced filtering, and strict filtering (shown in Algorithm \ref{alg:merge_points}).

\begin{algorithm}[htbp]
\caption{Merge Points Based on Distance Threshold}
\label{alg:merge_points}
\begin{algorithmic}[1]
\Require Dataset $D = \{\text{surveyId}, \text{lat}, \text{lon}, \text{speciesIds}\}$, distance threshold $T$
\Ensure Merged dataset $D_{merged}$
\State Convert latitude and longitude to radians
\State Sort dataset $D$ by the count of species in descending order
\State Build BallTree spatial index using haversine distance
\State Initialize an empty list $D_{merged}$
\For{each point $p_i$ in dataset $D$}
    \State Query neighbors within radius $\frac{T}{6371}$ using BallTree
    \State Initialize an empty set $S_{merged}$
    \State Compute local distance metrics $LAT_{KM}=111.4$, $LON_{KM}=111.32 \cdot \cos(lat_i)$
    \State Refine neighbors based on local latitude and longitude difference thresholds (0.32 km)
    \For{each neighbor $p_j$ within refined neighbors}
        \State Update $S_{merged} \leftarrow S_{merged} \cup \text{speciesIds}(p_j)$
    \EndFor
    \State Create representative point with surveyId, lat, lon, and merged speciesIds
    \State Append representative point to $D_{merged}$
\EndFor
\State \Return $D_{merged}$
\end{algorithmic}
\end{algorithm}

\subsection{Backbone Networks}
Our network is an improved version of the Tighnari model proposed in 2024.  
Given the increased training data, we upgraded the size of the Swin Transformer \cite{liu2021swin} backbone used for satellite image feature extraction from Tiny to Base. For the temporal modality, we retained the Temporal Swin-Transformer architecture from the Tighnari model. Specifically, when handling temporal cubes cropped to sizes $(4,18,12)$ and $(6,4,20)$, we set the patch sizes to $(3,3)$ and $(2,5)$, and the window sizes to $(3,2)$ and $(2,3)$, respectively. We stacked two Swin-Transformer stages: one with depth 2 and 12 attention heads, and another with depth 6 and 24 heads, for the two types of temporal cubes. This configuration of Swin-Transformer backbone showed highly competitive accuracy and stability in our ablation studies—only a ResNet18 with reduced input convolutional size could achieve comparable performance.

The effectiveness of Swin-Transformer-based models in feature extraction for plant species distribution prediction is well-supported. In such tasks, large color regions (low-frequency information) in images or temporal cubes are generally more important than textures and edges (high-frequency information). Thanks to its alternating window and shifted window attention, Swin Transformer naturally has a much larger receptive field than standard convolutions and is thus better at extracting low-frequency information. Moreover, its use of smaller patch sizes and local windowed attention, compared to ViT, avoids the shortcomings of global attention with large patches and enhances its ability to capture high-frequency details—such as environmental boundaries in satellite image patches or seasonal transitions in temporal cubes.

For tabular modality feature extraction, we replaced the original MLP backbone with the TabM network proposed by Yandex to enhance representational capacity. TabM is based on the BatchEnsemble method: on top of a single main model's weights, it introduces a pair of learnable rank-1 scaling factors for each sub-model, enabling parallel inference and ensemble-like output fusion with minimal extra parameters. TabM does not rely on attention mechanisms, yet achieves superior accuracy and generalization on standard benchmarks and open tabular data competitions compared to leading models such as TabNet (attention-based) and CatBoost (ensemble-based).

Furthermore, the neighborhood label aggregation-based modality is set as an optional input rather than a mandatory one. Since PO data contains severe label noise, using domain label aggregation as a feature would introduce further noise. This modality is only available when the training data consists solely of PA samples (shown in Algorithm \ref{alg:multimodal_swin}).
\begin{algorithm}[htbp]
\caption{Multimodal Ensemble with Swin Transformers}
\label{alg:multimodal_swin}
\begin{algorithmic}[1]
\Require Inputs: Table features $t$, Landsat series $x$, Bioclim series $y$, Sentinel image $z$
\Ensure Output: Class logits

\State \textbf{Tabular features:}
\State Extract features via Tab backbone: $t \leftarrow \text{TabBackbone}(t)$
\State Project and normalize: $t' \leftarrow \text{LayerNorm}(\text{MLP}(t))$

\State \textbf{Landsat features:}
\State Normalize input Landsat data: $x \leftarrow \text{LayerNorm}(x[:, :, :, :20])$
\State Extract features via SwinTransformer: $x \leftarrow \text{Swin}(x)$
\State Normalize features: $x' \leftarrow \text{LayerNorm}(x)$

\State \textbf{Bioclim features:}
\State Normalize input Bioclim data: $y \leftarrow \text{LayerNorm}(y[:, :, :18, :])$
\State Extract features via SwinTransformer: $y \leftarrow \text{Swin}(y)$
\State Normalize features: $y' \leftarrow \text{LayerNorm}(y)$

\State \textbf{Sentinel features:}
\State Extract Sentinel image features via SwinBase: $z \leftarrow \text{SwinBase}(z)$
\State Normalize features: $z' \leftarrow \text{LayerNorm}(z)$

\State \textbf{Concatenate features:} $concat\_vec \leftarrow [t', x', y', z']$
\State Fuse features with FusionHub (mode='tri\_serial'): $fused\_vec \leftarrow \text{FusionHub}(t', x', y', z')$

\If{$\text{dim}(concat\_vec) \neq \text{dim}(fused\_vec)$}
    \State Project $concat\_vec$ to match fusion dimension: $concat\_vec \leftarrow \text{Linear}(concat\_vec)$
\EndIf

\State Apply residual connection and dropout: $fused\_vec \leftarrow \text{Dropout}(fused\_vec + concat\_vec)$
\State Compute class logits: $logits \leftarrow \text{Linear}(fused\_vec)$
\Return $logits$
\end{algorithmic}
\end{algorithm}

\subsection{Modality Fusion}
We trained separate models for the three modalities(four inputs) and observed considerable differences in accuracy: the satellite image modality yielded the best performance, while the bioclimatic time series and tabular modalities lagged behind. This suggests that the modalities contribute unequally to overall performance, with a dominant modality often present.

In parallel cross-attention fusion, the module cannot recognize the dominant modality's prior importance, causing its contribution to be diluted by weaker modalities. Additionally, when each modality maintains independent sets of $K$, $Q$, and $V$ vectors, incompatibility between their $K$ and $V$ spaces complicates modality alignment, making attention weights less effective. The cross-attention implementation in the original Tighnari model also concatenated all attention representations at the output, preventing stacking of modules and thereby limiting the extraction of higher-order features.

To address these limitations, we designed a stackable serial three-modality cross-attention mechanism. First, we use independent linear layers to map the three modalities' inputs to a shared hidden dimension. Each modality generates its own queries ($Q$), while all concatenated modality features share a single key/value ($KV$) projection, greatly reducing parameter count and improving GPU utilization. This also unifies the $K$/$V$ vectors from different modalities into the same semantic space.

Unlike traditional cross-attention modules that compute attention in parallel for each modality, our design lets modality $A$ attend to the latest features of $B$ and $C$ (as keys/values) and normalizes its output; then, the updated $A$ and $C$ are used to update $B$; finally, the updated $A$ and $B$ are used to update $C$, forming an ordered iterative process. This approach allows the dominant modality to compute attention representations first, helping the model focus on key information, while serial multi-step attention updates help the module capture higher-order dependencies. Unlike previous designs, the outputs are not simply concatenated but are mapped back to the original dimension and combined with the input via residual connections, allowing the module to be stably stacked for extracting high-order features (shown in Algorithm \ref{alg:tri_modal_attention}) .

\begin{algorithm}[htbp]
\caption{Tri-Modal Cross Attention}
\label{alg:tri_modal_attention}
\begin{algorithmic}[1]
\Require Inputs: Modalities $X_a, X_b, X_c$; hidden dimension $D$; number of heads $h$
\Ensure Outputs: Updated modalities $X_a', X_b', X_c'$
\State Project inputs $X_a, X_b, X_c$ to hidden dimension $D$
\State Compute attention queries $Q_a, Q_b, Q_c$
\State Compute shared keys and values projections $KV$

\State Update $X_a$: Cross-attention with $X_b, X_c$
\State Update $X_b$: Cross-attention with updated $X_a$ and $X_c$
\State Update $X_c$: Cross-attention with updated $X_a$ and $X_b$

\State Apply FFN and normalization to updated modalities $X_a, X_b, X_c$
\State Project updated hidden representations back to original dimensions and add residual connections

\Return $X_a', X_b', X_c'$
\end{algorithmic}
\end{algorithm}

\subsection{Loss Function}
Exploratory Data Analysis (EDA) reveals that the mode of the number of species recorded per survey is 12, while the total number of species is 11,255. Thus, negative classes vastly outnumber positive classes. Traditional Binary Cross-Entropy (BCE) Loss treats positive and negative samples equally, which leads to overfitting on majority (negative) classes and insufficient learning for minority (positive) classes. Therefore, we aim to assign higher weights to difficult examples.

Moreover, negative labels are inherently noisy—an “absent” label does not guarantee that the species is truly absent. If the model correctly predicts a high probability of presence but is heavily penalized due to a mislabeled negative, effective feature learning is impeded. In addition, the geographical distribution of PA test samples differs from the training data, rendering Threshold Top-K methods ineffective due to their reliance on identical distributions between training and test sets. In such scenarios, fixing the binary classification threshold at 0.5 becomes particularly important, as it provides a stable and reproducible standard for positive/negative decision, independent of sample distribution.

Ultimately, we adopted Asymmetric Loss (ASL) \cite{ridnik2021asymmetric} as a replacement for BCE. The ASL formula is as follows:

\begin{equation}
\mathcal{L}_{\text{ASL}}
= \frac{1}{N} \sum_{i=1}^{N}
\bigl[
  -\, y_i \,(1 - p_i)^{\gamma_+} \log p_i
  \;-\;
  (1 - y_i)\,
  \max\!\bigl(p_i - m,\,0\bigr)^{\gamma_-}
  \log(1 - p_i)
\bigr],
\label{eq:asl_multi}
\end{equation}

Here, $N$ denotes the total number of labels; $y_i \in \{0,1\}$ is the ground-truth for the $i$-th label; $p_i \in [0,1]$ is the predicted probability that the $i$-th label is positive; $\gamma_+$ and $\gamma_-$ are the focusing parameters for positive and negative samples, respectively; and $m$ is the clipping threshold applied to negative samples.

ASL introduces several improvements over BCE:
\begin{itemize}
    \item \textbf{Asymmetric focusing:} Different focusing parameters ($\gamma_+$ for positives, $\gamma_-$ for negatives) adjust the contribution of hard/easy examples for each class separately.
    \item \textbf{Negative sample clipping:} The negative loss term is suppressed when the predicted probability $p$ is below the threshold $m$, which reduces the impact of potentially noisy negative labels on the gradient.
    \item \textbf{Robustness:} By mitigating the dominance of noisy negatives, ASL improves the model's robustness and overall feature learning capability.
\end{itemize}
\subsection{Training Strategy}
In this section we describe the training strategy for the two expert models in the MoE architecture.
We trained two separate models: one for inference on test samples geographically close to PA training samples (referred to as the in-distribution test set), and another for test samples in regions without any nearby PA training samples (the out-of-distribution test set). For in-distribution inference, we followed the same training procedures and post-processing strategies as the Tighnari model proposed in 2024.

For out-of-distribution inference, we adopted a two-stage training strategy. In the first stage, we mixed PA training samples with pseudo-labeled PO samples and pre-trained the model for three epochs with a relatively high learning rate (0.0001). This enabled the model to encounter and learn associations for many species present only in PO data. In the second stage, we fine-tuned the model for five additional epochs with clean PA data and a lower learning rate, further enhancing the model’s predictive capability for species in the PA set, while minimizing catastrophic forgetting for species only present in PO data. The number of epochs was determined by monitoring overfitting on PO data (as training set) while validating on PA data: since PO labels are much noisier, the model tends to overfit negatives faster. If PA and PO were trained together from the beginning, persistent loss reduction on the PA validation set would obscure the point when the PO set begins to overfit.

Overall, this training strategy balances rare species recognition with robust generalization on high-quality labeled samples.

\subsection{Post-Processing Strategy}
We implemented a divide-and-conquer inference process based on whether the test sample is geographically close to any PA training sample, separating the in-distribution and out-of-distribution test sets. The post-processing steps are adapted from the Tighnari model.

First, we used the Threshold Top-K method to select predicted species, then merged these predictions with high-probability species lists from geographically adjacent samples, deduplicating the results. For the in-distribution test model, the threshold and $K$ parameters in Threshold Top-K can be directly optimized via grid search based on validation set performance. We further selected the five nearest PA training samples to each test point, counting as present any species with over 80\% observed frequency.

For the out-of-distribution model, determining the optimal threshold for Threshold Top-K was less straightforward. With the help of the ASL loss, we controlled the optimal threshold to around 0.45--0.5, and further tuned this value using Kaggle’s submission feedback. Afterward, we selected the six closest PO training samples (obtained using the strict pseudo-label filtering strategy), and considered any species present in more than 50\% of them to adjust the model’s predictions.

\section{Experiments}
\subsection{Metrics}
To demonstrate the optimality of our chosen backbone networks, we conducted extensive comparative experiments. We used the weighted F1 Score and Binary Cross-Entropy (BCE) Loss as the evaluation metrics. The formulas are as follows:

For each PA test sample, let $Y_i$ denote the ground-truth set of species (as species IDs), and $\hat{Y}_i$ denote the predicted set. For every test sample, the submitted prediction must provide a list of predicted species. The micro $F_1$-score is then computed as:

\begin{equation}
F_1 = \frac{1}{N} \sum_{i=1}^{N} \frac{\mathrm{TP}_i}{\mathrm{TP}_i + \left( \mathrm{FP}_i + \mathrm{FN}_i \right)/2},
\label{eq:micro_f1}
\end{equation}

where
\begin{align*}
\mathrm{TP}_i & = \text{Number of predicted species truly present, i.e., } |\hat{Y}_i \cap Y_i|, \\
\mathrm{FP}_i & = \text{Number of species predicted but absent, i.e., } |\hat{Y}_i \setminus Y_i| ,\\
\mathrm{FN}_i & = \text{Number of species not predicted but present, i.e., } |Y_i \setminus \hat{Y}_i|.
\end{align*}

$N$ is the total number of test samples. This metric averages the per-sample $F_1$-score over all samples.

The Binary Cross-Entropy (BCE) Loss is defined as:

\begin{equation}
\mathcal{L}_{\text{BCE}} = -\frac{1}{N} \sum_{i=1}^{N} \left[ y_i \log(p_i) + (1 - y_i) \log(1 - p_i) \right],
\label{eq:bce_loss}
\end{equation}

where $N$ is the number of samples, $y_i \in \{0,1\}$ is the ground-truth label, and $p_i$ is the predicted probability for sample $i$.

The model was trained for approximately 10 hours on an H20 GPU for each run.

\subsection{Comparative Experiments}
Our comparative experiments demonstrate the superiority of our selected backbone networks and modality fusion approach. 
In the comparative experiments, we adopted the Tighnari model from 2024 as our base model, in which satellite image features were extracted using Swin Transformer Tiny, while the time-series cubes features were extracted by a modified Swin Transformer with adjusted hyperparameters such as patch size, window size, and depth. Tabular features were extracted using fully-connected layers.

We used the PO dataset merged through the strategy described above as the training set, and all PA samples (the 2024 PA training set) as the validation set, to evaluate the capability of various architectures to learn effectively from label-noisy data.

In Table 1, we fixed the time-series feature extraction network and experimented by replacing image extraction networks of different types and sizes. Through extensive comparisons, we found that the Swin Transformer Base significantly improved performance (shown in Table \ref{table1}).

In Table 2, we fixed the Swin Transformer Base as the satellite image feature extraction network and tested several visually-based backbone networks with hyperparameter tuning for extracting features from the time-series cubes. We found that our Temporal Swin Transformer, designed in 2024, achieved the best performance (shown in Table \ref{table2}).

In Table 3, we compared direct feature concatenation, two other stackable cross-attention methods, and our proposed sequential attention. We ultimately found that our proposed cross-attention mechanism significantly enhanced performance (shown in Table \ref{table3}).

%Due to time constraints, more comprehensive comparative experiments will be supplemented in the near future.

\begin{table}[htbp]
    \centering
    \caption{Comparison of Backbone Networks on Validation Set}
    \label{table1}
    \begin{tabular}{l l c c}
        \toprule
        Time series cube & Satellite image & validation & W\_F1 \\
        \midrule
        Swin & Dinosmall        & 0.00677 & 0.19715 \\
        Swin & Dinobase         & 0.00683 & 0.20017 \\
        Swin & Swintiny         & 0.00676 & 0.19653 \\
        \textbf{Swin} & \textbf{Swinbase}         & \textbf{0.00665} & \textbf{0.21705} \\
        Swin & Swinv2tiny       & 0.00672 & 0.19822 \\
        Swin & Swinv2base       & 0.00678 & 0.20874 \\
        Swin & ConvNext\_base   & 0.00673 & 0.21286 \\
        Swin & ConvNext\_tiny   & 0.00679 & 0.20401 \\
        Swin & ConvNextv2\_base & 0.00685 & 0.19819 \\
        Swin & ConvNextv2\_tiny & 0.00672 & 0.20562 \\
        Swin & vit\_base        & 0.00680 & 0.19444 \\
        Swin & Efficient\_b2    & 0.00675 & 0.19069 \\
        Swin & Efficient\_b5    & 0.00668 & 0.20945 \\
        Swin & Efficient\_b0    & 0.00667 & 0.20715 \\
        \midrule
        ViT  & Swinbase         & 0.00679 & 0.19889 \\
        ViT  & Swintiny         & 0.00682 & 0.20393 \\
        ViT  & vitbase          & 0.00686 & 0.19559 \\
        ViT  & ConvNextbase     & 0.00671 & 0.19768 \\
        ViT  & ConvNexttiny     & 0.00683 & 0.19112 \\
        ViT  & ConvNextv2base   & 0.00677 & 0.20114 \\
        ViT  & ConvNextv2tiny   & 0.00675 & 0.18632 \\
        ViT  & Dinosmall        & 0.00682 & 0.18018 \\
        ViT  & Swinv2base       & 0.00696 & 0.18047 \\
        ViT  & Swinv2tiny       & 0.00674 & 0.19720 \\
        ViT  & Efficient\_b0    & 0.00688 & 0.19214 \\
        ViT  & Efficient\_b2    & 0.00690 & 0.17879 \\
        ViT  & Efficient\_b5    & 0.00672 & 0.19483 \\
        \bottomrule
    \end{tabular}
\end{table}

\begin{table}[htbp]
    \centering
    \caption{Performance Comparison of Backbone Combinations (Swinbase as Satellite Image Backbone)}
    \label{table2}
    \begin{tabular}{l l c c}
        \toprule
        Time series & Satellite image & validation & W\_F1 \\
        \midrule
        Resnet50      & Swinbase & 0.00685 & 0.20320 \\
        Resnet34      & Swinbase & 0.00664 & 0.20592 \\
        Resnet18      & Swinbase & 0.00672 & 0.21323 \\
        convtiny      & Swinbase & 0.00673 & 0.20865 \\
        Efficient\_b2 & Swinbase & 0.00680 & 0.18558 \\
        Efficient\_b0 & Swinbase & 0.00687 & 0.19104 \\
         ViT  & Swinbase         & 0.00679 & 0.19889 \\
        \textbf{Swin}          & \textbf{Swinbase} & \textbf{0.00665} & \textbf{0.21705} \\
        \bottomrule
    \end{tabular}
\end{table}

\begin{table}[htbp]
    \centering
    \caption{Comparison of Different Fusion Methods}
    \label{table3}
    \begin{tabular}{l c c}
        \toprule
        Fusion Method & validation & W\_F1 \\
        \midrule
        concat & 0.00689 & 0.20306 \\
        pair   & 0.00675 & 0.20535 \\
        four   & 0.00683 & 0.21268 \\
        \textbf{serial} & \textbf{0.00665} & \textbf{0.21705} \\
        \bottomrule
    \end{tabular}
\end{table}

\subsection{Ablation Studies}
Additionally, to demonstrate that the MoE approach can effectively enhance the model's performance on OOD datasets, we conducted an ablation study using the 2024 leaderboard (where test samples and PA training samples are in-distribution) and the 2025 leaderboard (with OOD samples). Specifically, we tested three scenarios: using only PA data, using both PA and PO data to train a single model, and our previously mentioned MoE model.

The results show that the MoE model achieved the best scores on both the 2024 and 2025 test sets. Moreover, the performance improvement brought by the MoE model was particularly prominent on datasets containing OOD samples (shown in Table \ref{table4}).
\begin{table}[htbp]
    \centering
    \caption{Comparison of Scores in GeoLifeCLEF 2024 and 2025}
    \label{table4}
    \begin{tabular}{lcccc}
        \toprule
        & \multicolumn{2}{c}{2024 GeoLifeCLEF} & \multicolumn{2}{c}{2025 GeoLifeCLEF} \\
        \cmidrule(r){2-3}\cmidrule(r){4-5}
        & Private Score & Public Score & Private Score & Public Score \\
        \midrule
        PA Only & 0.36501 & 0.36541 & 0.17290 & 0.20604 \\
        PA + PO & 0.33335 & 0.33597 & 0.19107 & 0.21860 \\
        MoE & \textbf{0.36908} & \textbf{0.37246} & \textbf{0.21689} & \textbf{0.24493} \\
        \bottomrule
    \end{tabular}
    \label{tab:GeoLifeCLEF_scores}
\end{table}

\section{Conclusion}

\subsection{Limitations and Reflections}
In our framework, pseudo-labels are generated based on prior rules, resulting in the number of pseudo-labeled species being less than or equal to the actual number of species. Moreover, no semi-supervised process is applied for incremental iteration in later stages.

We recognize that PO data inherently contains label noise that is class-dependent; that is, some species labeled as negative samples may in fact have been present during the corresponding spatio-temporal survey. However, our current work does not incorporate strategies to explicitly model or mitigate such label noise.

Our species distribution model is essentially a multi-label binary classification task. The large number of labels leads to overly generalized feature extraction by the backbone network, negatively impacting model performance. In addition, the vast majority of labels only appear in a few samples (i.e., a long-tailed distribution), making the model prone to overfitting on the few common labels while ignoring the rare ones.

\subsection{Future Work}
Visualization results indicate significant differences in the number of species observed per survey. Instead of using a fixed $K$ as in the Threshold Top-K method, we propose allowing the model to predict the appropriate $K$ for each sample based on tabular features.

For binary classification of each species, we plan to anchor positive labels as immutable based on prior knowledge, while allowing negative labels to transition to positives with a certain probability. Specifically, we intend to estimate the probability of negative-to-positive transitions in a transition matrix and use this matrix to down-weight the loss penalty for misclassified negatives.

Furthermore, we aim to exploit inter-label correlations by constructing a graph structure, treating each species label as a node and connecting nodes that co-occur in the same observation. After traversing all samples, this yields a species co-occurrence graph. Based on this graph, we can apply graph clustering to partition species nodes, or use the structure to initialize a Graph Neural Network (GNN) layer at the output stage, thereby leveraging label co-occurrence as a prior to refine predictions.

\begin{acknowledgments}
The data for this paper is organized and published by INRIA. We express our gratitude to all the institutions and individuals involved in data collection and processing, including but not limited to the Global Biodiversity Information Facility (GBIF, www.gbif.org), NASA, Soilgrids, and the Ecodatacube platform. We also express our sincere appreciation to the LifeCLEF and GeoLifeCLEF teams for organizing the challenges and providing timely support.
All authors contribute helpful ideas during the course of the competition and participate in writing and revising the paper, so all authors are co-first authors. All of the authors, as corresponding authors, are obliged to reply to emails to provide readers with the relevant code and data of this work and explain the details of the work. Among them, Haixu Liu, as the first corresponding author, is responsible for the necessary communication for the publication of the article.
\end{acknowledgments}

%% The declaration on generative AI comes in effect
%% https://ceur-ws.org/GenAI/Policy.html
\section*{Declaration on Generative AI}
 During the preparation of this work, the authors used ChatGPT-4.5 and ChatGPT-o3 in order to: Text Translation and Formatting assistance. After using these tools/services, the authors reviewed and edited the content as needed and take full responsibility for the publication’s content. 

%%
%% Define the bibliography file to be used
\bibliography{sample-ceur}

\appendix

\section{Online Resources}

Code in Colab
\begin{itemize}
\item \href{https://drive.google.com/file/d/1XzPQQ2bLucriDTaGzcPyjwdRNKsUeGQd/view?usp=sharing}{2024 GeoLifeCLEF},
\item \href{https://colab.research.google.com/drive/1-Mx-fC15T-pRCI8D1f93aljfAt8vTnG7?usp=sharing}{2025 GeoLifeCLEF}.
\end{itemize}

\noindent Dataset in Kaggle
\begin{itemize}
\item \href{https://www.kaggle.com/competitions/geolifeclef-2025/data}{GLC25 PA and PO Tabular},
\item \href{https://www.kaggle.com/datasets/liuhaixu/glc25-po-image}{GLC25 PO Satellite Image},
% \item \href{https://www.overleaf.com/project/5e76702c4acae70001d3bc87}{Overleaf},
\item
\href{https://www.kaggle.com/datasets/liuhaixu/glc25-po-cube}{GLC25 PO Time Series Cube},
\item
\href{https://www.kaggle.com/datasets/liuhaixu/glc25-po-bioclimatic}{GLC25 PO Bioclimatic},
\item
\href{https://www.kaggle.com/models/liuhaixu/4ch_timm}{Four Channel Timm Backbone}.

\end{itemize}

\end{document}